# Parkinson's Disease Diagnosis Through Deep Learning: A Novel LSTM-Based Approach for Freezing of Gait Detection

Aqib Nazir Mir[1], Iqra Nissar, Mumtaz Ahmed[1], Sarfaraz Masood[1], Danish Raza Rizvi[1]

[1]Department of Computer Engineering, Jamia Millia Islamia, New Delhi, 110025, India

**Abstract.** Deep learning holds tremendous potential in healthcare for uncovering hidden patterns within extensive clinical datasets, aiding in the diagnosis of various diseases. Parkinson's disease (PD) is a neurodegenerative condition characterized by the deterioration of brain function. In the initial stages of PD, automatic diagnosis poses a challenge due to the similarity in behaviour between individuals with PD and those who are healthy. Our objective is to propose an effective model that can aid in the early detection of Parkinson's disease. We employed the VGRF gait signal dataset sourced from Physionet for the purpose of distinguishing between healthy individuals and those diagnosed with Parkinson's disease. This paper introduces a novel deep learning architecture based on LSTM network for automatically detecting freezing of gait episodes in Parkinson's disease patients. In contrast to conventional machine learning algorithms, this method eliminates manual feature engineering and proficiently captures prolonged temporal dependencies in gait patterns, thereby improving the diagnosis of Parkinson's disease. The LSTM network resolves the issue of vanishing gradients by employing memory blocks in place of self-connected hidden units, allowing for optimal information assimilation. To prevent overfitting, dropout and L2 regularization techniques have been employed. Additionally, the stochastic gradient-based optimizer, Adam, is used for the optimization process. The results indicate that our proposed approach surpasses current state-of-the-art models in FOG episode detection, achieving an accuracy of 97.71%, sensitivity of 99%, precision of 98%, and specificity of 96%. This demonstrates its potential as a superior classification method for Parkinson's disease detection.

## 1 Introduction

Parkinson's disease occurs due to the shortage of dopamine neurons in the specific part of the brain called as substantia nigra. The statistics from the Parkinson's Foundation shows that approximately one million individuals were affected by PD in the United States in 2020 [1]. PD diagnosis predominantly relies on motor symptoms, encompassing gait issues, tremors, rigidity, etc. Although therapy and medications are presently employed to manage these symptoms as presently there is no known cure available but initial diagnosis of PD can aid in planning appropriate therapies or medications to alleviate disease progression. Currently, in medical sciences, the diagnosis of PD is primarily relied upon the observations determined by clinicians and these observations may sometimes lead to false positives. Therefore, there is a growing shift towards incorporating deep learning (DL) algorithms in PD diagnosis. DL algorithms are increasingly valuable in bioinformatics due to their capacity to handle extensive datasets and uncover biomarkers, which not only results in more accurate predictions but also reduces the diagnosis time.

Various physiological signals including gait patterns, voice signals, handwritten samples and tremor are employed in Parkinson's diagnosis. Speech analysis can reveal non-motor symptoms of PD, with sustained vowel and word phonation aiding in differentiation between normal and PD affected subjects [2–4] . The potential biomarker in handwriting involves kinesthetic elements, cognitive, and perceptual motor. This involves tests such as the static spiral test, stability test and dynamic spiral test using a graphics tablet and LCD monitor. Notable ML results in this area have been reported in [5–8].

ML models have also been crucial in analyzing complex genetic and transcriptomic data to understand the pathology of PD [9–12]. Brain imaging modalities, particularly structural magnetic resonance imaging, provide valuable insights for computer-based PD diagnosis due to their high-resolution imaging of brain tissues. Additionally, techniques like DatScan are also used for PD diagnosis. ML algorithms have been employed to detect dopamine presence in the brain using MRI images [13–15].

Gait-based classification for diagnosing motor symptoms deficits is both cost-effective and easily obtainable [16]. Notably, the gait cycle exhibits distinctive features such as deterministic behavior, periodicity, and spatio-temporal characteristics. Unlike methods relying on speech or handwriting, which primarily address non-motor functions, gait analysis allows to evaluate motor functions and helps to assess the severity of the subjects' condition [17]. Consequently, clinicians can recommend appropriate therapeutic interventions to slow down disease progression. Recently, significant progress has been made in utilizing deep learning algorithms to predict PD stages through gait analysis. The contributions of this work are summarized as:
  i. The work focuses on solving the binary class classification problem in Parkinson's disease by employing an LSTM classifier that effectively leverages temporal information within gait sequences.
  ii. To prevent overfitting of data, this research employs dropout as well as L2 regularization techniques, ensuring the model generalizability.
  iii. The Adam optimizer, known for its reduced memory requirements and minimal hyperparameter tuning, is used for optimization enhancing efficiency in model training.

The paper is organized as: section 2 discusses the different works that have been accomplished for freezing of gait detection. In section 3, the proposed methodology has been presented. Results and discussion have been entailed in section 4. Additionally, the comparative analysis has also been presented. Section 5 concludes the paper.

## 2 Related Work

Alharthi et al.[18] introduced an interpretable end-to-end deep learning system designed to combine raw gait signals, leading to the development of a PD detection- classification model. This model achieved an average of 95.5% for F1 score with a low 0.28% standard error. To gain insights into the decision-making process of the model, Layer-wise Relevance Propagation (LRP) was employed for interpretation.

This paper [19] tackles the challenge of assessing PD motor impairments using the MDS-UPDRS scale. The authors proposed a novel ordinal focal network for estimating MDS-UPDRS scores. They also proposed a regularization method called rater confusion estimation (RCE) to handle inter-rater variabilities. Their approach was applied to gait and finger tapping scores from video recordings. Results on a clinical dataset showed 72% classification accuracy with majority vote ground truth and 84% accuracy when predicting at least one rater's score. This highlights the potential of computer-assisted technology for tracking PD patients motor impairments, even in cases of disagreement among clinical experts.

In this study [20], the primary focus was on drug-induced Parkinsonism in older adults. The researchers employed ST-GCN to forecast clinical scores pertaining to parkinsonism using video data. They conducted a comparative analysis between ST-GCN models and conventional regression models, as well as temporal convolutional network baselines. This involved extracting joint trajectories from video with the aid of different pose estimation libraries and the Microsoft Kinect device. The findings consistently demonstrated that proposed model utilizing 3D joint trajectories outperformed other methodologies. Nevertheless, predicting Parkinsonism scores in new subjects remains a challenge, with the most effective models achieving F1-scores of $0.40 \pm 0.02$ for SAS-gait and $0.53 \pm 0.03$ for UPDRS.

This paper [21] emphasized the profound impact of deep learning in healthcare, particularly its role in uncovering hidden patterns within clinical data for disease detection, with a focus on monitoring Parkinson's disease (PD) using wearable leg sensors. The primary objective was to detect PD symptoms, particularly FOG, which varies in severity among patients. The paper introduced a patient-specific approach, employing a LSTM network for FOG detection, and compares it with conventional algorithms like linear SVM using sensor data. The results highlight the LSTM's superiority, achieving an average accuracy of 83.38%, compared to SVM's accuracy 79.48%. This underscores the effectiveness of deep learning, specifically LSTM networks, in identifying patient-specific PD symptoms using wearable sensor data.

This research [22] introduced a novel technique for differentiating among Parkinson affected tremor (PT) and normal subjects essential tremor (ET) by examining tremors in both postural and resting states. The method entails capturing hand tremors in both positions and employing a convolutional LSTM network to acquire the ability to distinguish between these tremors. The study collected 3D landmark points of hand movements using a Leap Motion Controller, pre-processed the tremor data, and trained the convolutional LSTM for classification. Evaluation was performed on a dataset comprising 40 subjects, including 23 with PT and 17 with ET. The results demonstrated that the accuracy of classification improves significantly when considering both resting and postural tremors, reaching an impressive accuracy rate of 90% for distinguishing between the two types of tremors.

This paper [23] introduced an innovative PD detection system using DL, specifically 1D CNN (1D-Convnet) within a DNN classifier. The network consists of two components: the initial section incorporates 18 parallel 1D-Convnet layers that correspond to the input signals, while the subsequent part constitutes a fully connected network responsible for combining their results to facilitate classification. The system achieves an impressive 85.3% accuracy in predicting PD severity based on the UPDRS, marking the first algorithm capable of UPDRS-based severity prediction.

Veeragawan et al. [24] introduced a non-invasive approach to diagnose and evaluate the severity of PD in its early and moderate stages by VGRF data. The method involves extracting gait-related features from the VGRF data, which are then used to train an ANN model. The results showed an impressive accuracy of 87.1% in assessing the severity of PD.

## 3 Methodology

Fig. 1 outlines the proposed workflow for the detection of Freezing of Gait (FOG) in Parkinson's patients. The comprehensive exploration of the sequential steps involved are discussed below. Additionally, Fig. 2 provides a detailed depiction of the algorithm.

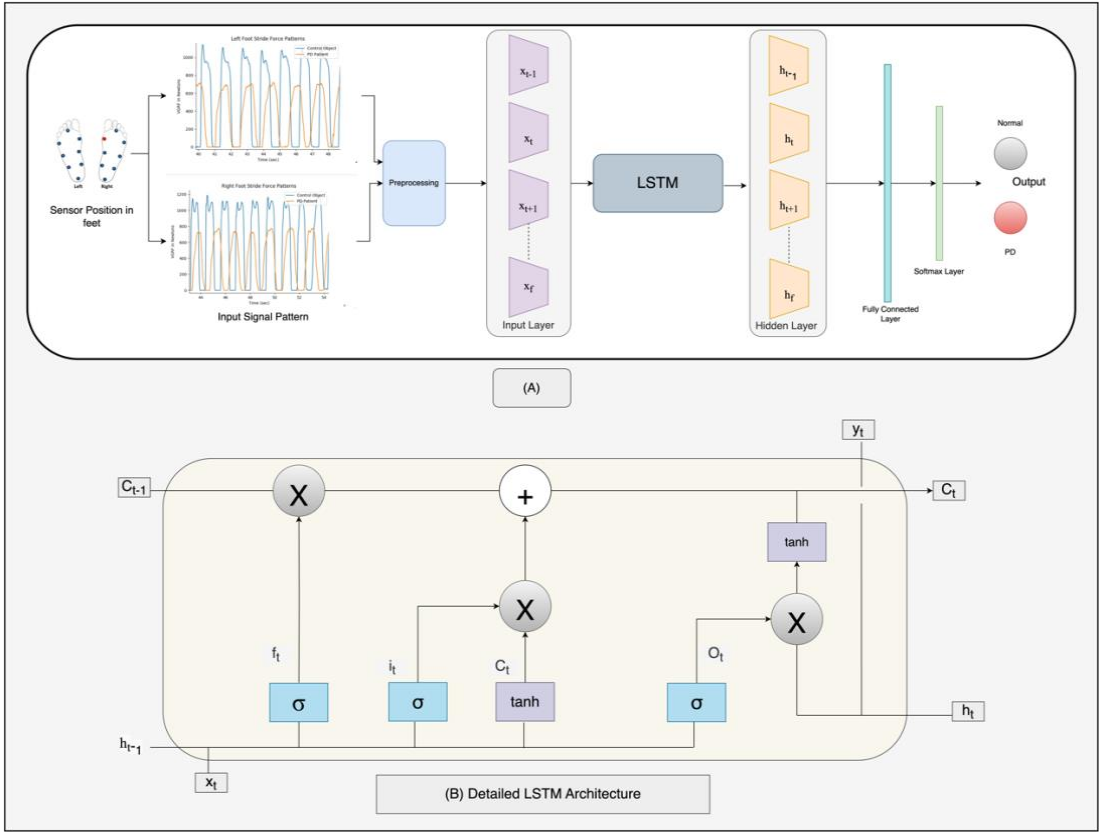

**Fig. 1.** (A): Proposed workflow for FOG detection in Parkinson's subjects. (B): Detailed LSTM architecture.

### 3.1 Dataset

The proposed model is evaluated using a PD dataset comprising sensor signals, retrieved from PhysioNet [25]. This dataset comprises gait signals obtained from a total of 166 individuals, with 93 being diagnosed with PD and 73 classified as healthy subjects. Eight sensors were positioned under the subject's foot to obtain the vertical ground reaction force (VGRF) and 100 samples were recorded in one second and subsequently digitized. Table 1 entails a description of the data format. Upon examining the time series data from the subjects, distinctive variations in walking patterns between PD and control individuals become evident, the analysis of which is visually presented in Fig. 3.

### 3.2 Data Pre-processing

To ensure uniformity in input data length, each sample consists of 19 attributes of data with varying column lengths due to different recording durations for subjects (ranging from 1000 to 12119 frames). To address this, the datasets were partitioned into equal-sized segments of 500 frames. This division allowed for long recordings to be segmented into multiple chunks, maintaining consistency. The timestamp columns, which did not provide relevant gait information, were removed. Consequently, the size of the final sample comprises 18 attributes and 500 instances ($x_i = [x_1 ... x_{18}] \in \mathbb{R}^{500*18}$).

**Table 1.** Dataset description.

| Input column | Variable | description |
|---|---|---|
| 1 | $t_n$ | Time (s). |
| 2-9 | $ln1, ln2, ....., ln8$ | "left foot VGRF records for eight sensors" |
| 10-17 | $rn1, rn2, ......, rn8$ | "right foot VGRF records for eight sensors" |
| 18 | $C_{18}$ | "left foot total force" |
| 19 | $C_{19}$ | "right foot total force" |

---

<u>Algorithm: Parkinson's Disease Prediction using LSTM</u>
1. Input:- d-dimensional gait pattern x, Trained LSTM model
2. Output:- Prediction of Parkinson's disease
3. Initialization:
   Number of gait signals (No. of dimensions) = 18
   Number of classes (C) = 2
   Signal dimension (d) = 12000
   Number of segments (M) = 500
4. Procedure PD_LSTM(x, C, LSTM):
5. Select the gait segment length (L).
6. Partition the VGRF signals into M segments of size 500x1.
7. Initialize LSTM states:
   $h_{t-1} = 0$ (previous hidden state)
   $C_{t-1} = 0$ (previous cell state)
8. for i from 1 to M do
9. Extract the $i^{th}$ segment: $x_t$
10. Perform LSTM calculations:
    $i_t = \sigma(W_{hi}h_{t-1} + W_{xi}x_t + b_i)$  # Input gate
    $f_t = \sigma(W_{hf}h_{t-1} + W_{xf}x_t + b_f)$  # Forget gate
    $C_t = f_t \odot C_{t-1} + i_t \odot C'_t$  # Cell state
    $o_t = \sigma(W_{ho}h_{t-1} + W_{xo}x_t + b_o)$  # Output gate
    $o_t = h_t \odot \tanh(C_t)$  # LSTM Output
11. Apply a fully connected layer to get $s_t$: $s_t = h_t(y_t)$  # FC Layer
12. Store the output state $h_t$ for the next iteration.
13. Perform average pooling over the output states:
14. $E = AP(S_t, S_{t-1}, ....S_{t-M})$;
15. Determine the predicted class(PC):
16. IHP = argmax(PC)  # Index of the highest probability
17. Set the predicted class as $\hat{Y}$: $\hat{Y}$ = IHP
18. Return the predicted class $\hat{Y}$ as the output
19. End of procedure

**Fig. 2.** Implementation of LSTM model for FOG detection.

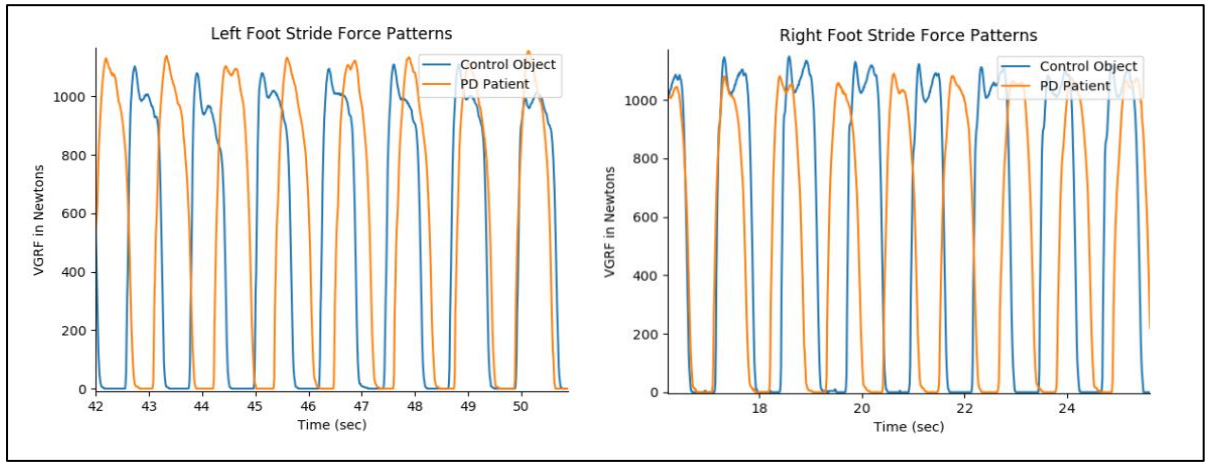

**Fig. 3.** Signal samples showing variations of walking patterns of control and PD subjects.

## 3.3 LSTM Architecture

Hochreiter and Schmidhuber introduced LSTM to address the limitations of vanilla RNN, specifically the exploding and vanishing gradient problems. LSTM excels in learning long-term dependencies in sequential data [26,27]. In contrast to RNN, LSTM employs memory cells in lieu of conventional nodes within the hidden layer. This allows for the retention and recall of information over prolonged durations. Fig. 1 (B) provides a visual representation of the LSTM architecture. The memory block is composed of three gates. These are the input gate, forget gate, and output gate. Multiplicative gate units are used to minimize the impact of irrelevant inputs [28]. Retention of information from previous state is decided by the forget gate and is calculated as follows:

$$f_t = \sigma(W_{hf} h_{t-1} + W_{xf} x_t + b_f) \tag{1}$$

Where $h_{t-1}$ is the previous block output, $x_t$ is the input sequence and b is the bias vector; $W_{xf}$ and $W_{hf}$ denote the weight matrices corresponding to the current cell's input vector and the preceding cell's output vector, respectively. The input layer is responsible for selecting which information gets stored based on the present input vector. The output for the current time step is calculated by the output gate. The equations are defined as:

$$i_t = \sigma(W_{hi} h_{t-1} + W_{xi} x_t + b_i) \tag{2}$$
$$o_t = \sigma(W_{ho} h_{t-1} + W_{xo} x_t + b_o) \tag{3}$$

Combining the forget and input gate, the equation for the current cell state is given as:

$$C_t = f_t \odot C_{t-1} + i_t \odot C'_t \tag{4}$$

Where, at time step t $C_t$ = cell state and $\odot$ = element-wise multiplication. $f_t \odot C_{t-1}$ and $i_t \odot C'_t$ determine the information taken from the preceding cell state and current input [34]. Using the tanh activation function, the $C'_t$ is calculated as:

$$C'_t = \tanh(W_{hC} h_{t-1} + W_{xC} x_t + b_C) \tag{5}$$

The value of the hidden state is calculated as:

$$o_t = h_t \odot \tanh(C_t) \tag{6}$$

### 3.3.1 Activation Function-ReLU

ReLU is a widely employed activation function in deep neural networks. It is favored for its parameter-free, non-saturating nature, which aids in expediting the convergence of stochastic gradient descent (SGD) [29]. In contrast to saturated activation functions such as sigmoid and tanh, ReLU has shown notable improvements in LSTM performance. It achieves faster convergence and higher accuracy by maintaining positive values and nullifying negative inputs. This enables effective gradient propagation during training. Moreover, ReLU provides a straightforward operation and mitigates the vanishing gradient problem, as the derivative remains a constant value on the positive side. The graphical representation of ReLU function is depicted in Fig. 4.

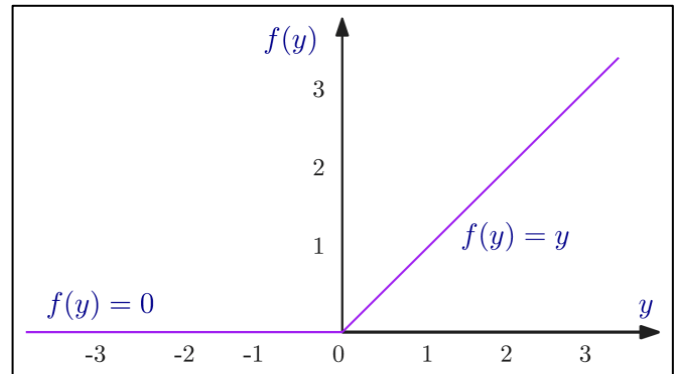

**Fig. 4.** Graphical representation of ReLU activation function.

### 3.3.2 L2 regularization

It is a technique that adjusts the loss function to impose penalties on excessive weight values, thereby reducing generalization error. This adjustment helps prevent the weights from becoming too large, which would make the deep neural network more sensitive to noise. The cost function with L2 regularization is given by:

$$\hat{l}(w, X) = l(w, X) + \lambda R(w) = l(w, X) + \lambda \sum i\, |w_i^2| \tag{7}$$

Here, $\lambda$ represents the regularization strength, $l(w, X)$ and $R(w)$ are cross-entropy loss function and a convex function respectively.

### 3.3.3 Dropout

Unlike L2 regularization, which penalizes large weight coefficients, dropout modifies the connections within the network by randomly excluding neurons with a probability of $p$. This approach aims to prevent co-adaptation among hidden nodes in the deep neural network, promoting more robust learning [30]. The core concept of dropout involves randomly deactivating portions of the model during each training iteration. The dropout function is expressed as:

$$q_j = \sum i\, W_{ij} V_i x_i + b_i \tag{8}$$

Here, $V_i$ = Bernoulli random variables independent vector. For $V_i$ = zero, its corresponding input node $x_i$ is excluded from the computation.

### 3.3.4 Adam optimizer

A combination of the RMSProp and momentum-based gradient descent methods known as adam optimizer and is a highly utilized approach for training deep neural networks [31,32]. It can effectively replace the standard SGD algorithm for updating the weights in a DNN. This method is memory-efficient, and it requires the tuning of only a few hyperparameters. Additionally, updates to the parameters ensure a bounded norm. As a result, Adam optimization finds extensive application across various deep neural network tasks [33].

### 3.3.5 Softmax layer

It is the last layer in the DNN which is responsible for determining the probability of the input which belongs to a particular class. During training, the cross-entropy loss function quantifies the variation between the true label p and predicted label q. To update the weights w and biases b, this function is employed as:

$$h(p, q) = -\sum i\, p(x) log q(x) \tag{9}$$

## 4 Results and discussion

The LSTM network was trained through a systematic process of adjusting various parameters to maximize performance during both training and testing phases. This experimentation involved multiple trials to evaluate the proposed model's performance. During the evaluation process, we refined our system by fine-tuning specific parameters like initial learning rate (LR), batch size, epoch count, and the configuration of hidden layers. The best-performing parameter values, as determined by our study, are summarized in Table 2. These parameters represent the configuration that yielded the highest accuracy and effectiveness for our model. For accessing the reliability of the model, precision, sensitivity, specificity and accuracy were employed.

A three-layer model (input layer, one hidden layer and output layer) is used, where the input layer maintains a fixed number of nodes. The node count in the hidden layer was changed from 64 nodes to 256 nodes, to find the ideal number of nodes for better performance. This is an essential component as it enables the model to learn complex features and patterns. Also, the number of epochs were altered from 50 to 80. The value for L2-regularization was taken as 0.0005, 0.005 and 0.0005. The different values of learning rate for three different models were 0.001, 0.0001 and 0.0001 respectively. The batch size was taken as 128, 64 and 64. For all these different values for different models, different values of performance metrics were calculated. At the last layer, the softmax function is used. This function is suitable for binary classification tasks, as it assigns probabilities to each class. Adam and categorical cross entropy were used as the optimizer and the loss function respectively. The data split ratio was taken as 70:30 for training and validation respectively.

Table 2. Performance metrics for LSTM architecture with different hyperparameters.

| Model | epochs | $L_2$ Reg. | Initial LR | Batch size | Precision | Sensitivity | Specificity | Accuracy (%) |
|---|---|---|---|---|---|---|---|---|
| Model 1 LSTM (64 neurons) | 50 | 0.0005 | 0.001 | 128 | 0.81 | 0.80 | 0.82 | 76.92 |
| Model 2 LSTM(128 neurons) | 60 | 0.005 | 0.0001 | 64 | 0.90 | 0.88 | 0.89 | 88.94 |
| **Model 3 LSTM(256 neurons)** | **80** | **0.0005** | **0.0001** | **64** | **0.98** | **0.99** | **0.96** | **97.71** |

The training process is run for 50 epochs in Model 1, 60 epochs in Model 2, and 80 epochs in Model 3. This parameter determines how many times the entire dataset is used for training. L2 regularization is employed to prevent overfitting. In Model 1 and Model 3, a regularization strength of 0.0005 is used, while in Model 2, a higher value of 0.005 is chosen. The initial learning rates are set at 0.001, 0.0001, and 0.0001 for Model 1, 2, and 3 respectively. This parameter regulates the stride taken during optimization, while the batch size denotes the sample count processed. In Model 1, a batch size of 128 is used, while in Model 2 and Model 3, a smaller batch size of 64 is employed.

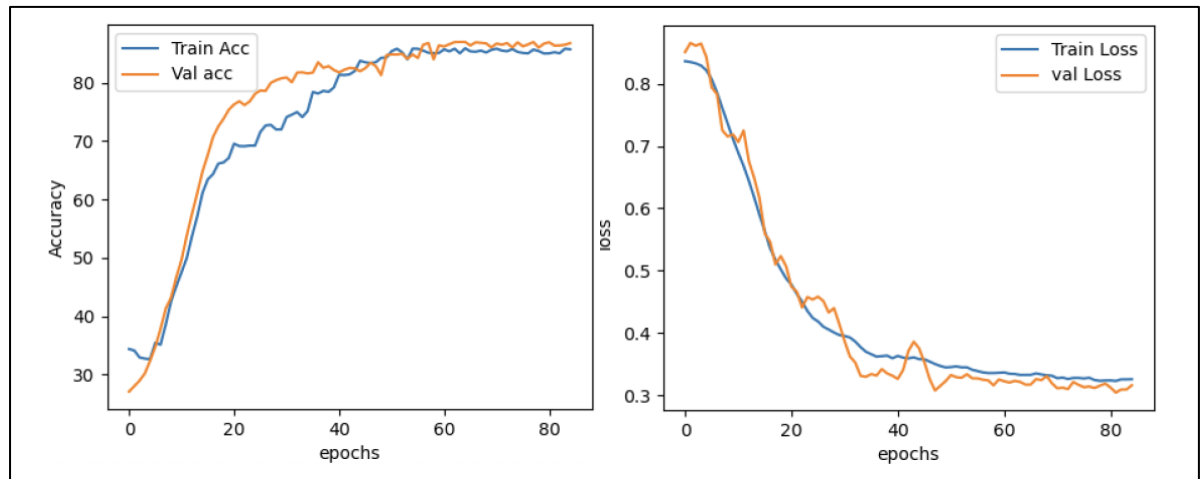

**Fig. 5.** Model training and validation curves for accuracy and loss parameters.

After conducting the experiments with different configurations, Model 1, which has 64 nodes in the hidden layer, 50 epochs, 0.001 learning rate, and a batch size of 128, yielded 0.81, 0.80, and 0.82 as precision, sensitivity, and specificity respectively. Also, an accuracy of 76.92% was reported. Model 2 has 128 nodes in the hidden layer, 60 epochs, 0.0001 learning rate, and a batch size of 64, yielded 0.90, 0.88, 0.89 as precision, sensitivity, specificity respectively with the accuracy of 88.94%. Model 3, which has 256 nodes in the hidden layer, 80 epochs, 0.0001 learning rate, and a batch size of 64, achieves the best performance. This model yields a precision of 0.98, sensitivity of 0.99, specificity of 0.96, and an impressive accuracy of 97.71%. These metrics show the model effectiveness in accurately classifying the data. Fig. 5 demonstrates the plots of accuracy and loss obtained while performing training and validation respectively.

### 4.1 Comparative analysis

In this section, an overview of different studies and their methodologies for detecting FOG episodes in individuals with PD has been presented in Table 3. Oktay et al [22] employed an LSTM-based approach for FOG detection, achieving a high sensitivity of 95.24%, indicating a low false negative rate. The model demonstrated a best accuracy of 90.0%. Tautan et al [34] used a 1D-CNN for their FOG detection. While the accuracy is not provided, the sensitivity and specificity metrics indicate that the model performs reasonably well in identifying positive cases (sensitivity=83.77%) and negative cases (specificity=81.78%).

**Table 3.** Results of the proposed LSTM model and the existing works on FOG detection.

| Author | Year | Methodology | Accuracy | Sensitivity | Specificity |
|---|---|---|---|---|---|
| Oktay et al [22] | 2020 | LSTM | 90.0% | 95.24 | 86.96 |
| Tautan et al.[34] | 2020 | 1D-CNN | NA | 83.77% | 81.78% |
| Mekruksavanich et al.[35] | 2021 | Squeeze and Excite CNN | 95.66% | 95.66% | NA |
| El-ziaat et al.[36] | 2022 | CONV-LSTM | 94.5% | 92.1% | 94.1% |
| Lin et al. [37] | 2023 | CNN+Majority voting | 83.0% | 85.4% | 82.7% |
| Hou et al. [38] | 2023 | Depth wise CNN | 85.0% | 81.0% | 88.0% |
| **Proposed Model** | **2023** | **LSTM** | **97.71%** | **99%** | **96%** |

Mekruksavanich et al [35] utilized a Squeeze and Excite CNN architecture, achieving high accuracy and sensitivity scores, both at 95.66%. Specificity is not reported. El-ziaat et al [36] employed a CONV-LSTM architecture for FOG detection, showing a high specificity of 94.1%. The model reported an accuracy and a sensitivity of 94.5% and 92.1% respectively. Lin et al [37] used a combination of CNN and a majority voting scheme for their FOG detection. The model obtained an accuracy of 83.0%, specificity of 82.7%. with a sensitivity of 85.4%. Hou et al [38] utilized a depth-wise CNN for FOG detection. The model reported an accuracy, specificity and sensitivity of 85.0%, 88.0%, 81.0% respectively.

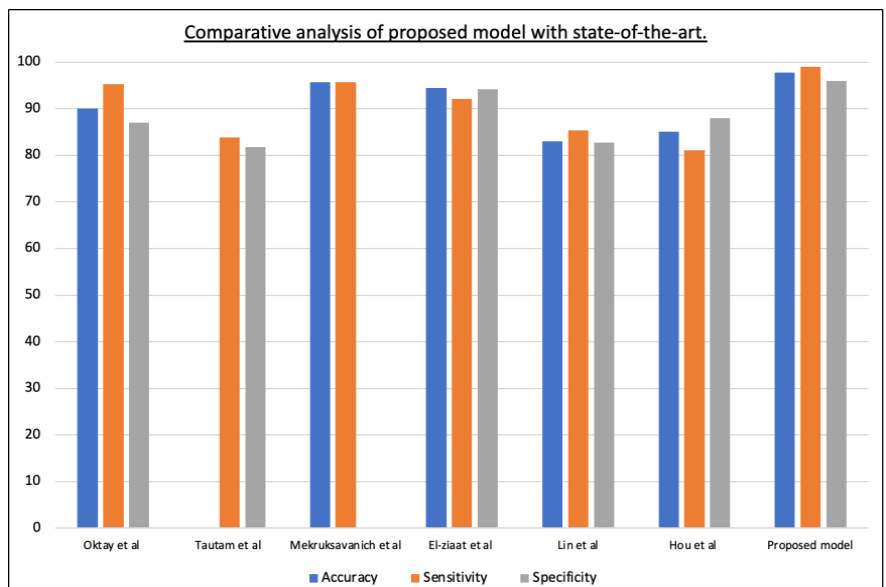

**Fig. 6.** Bar chart depicting the proposed LSTM model's performance with the SOTA.

The proposed model based on an LSTM architecture demonstrates exceptional performance, particularly in terms of sensitivity (99%), indicating an extremely low false negative rate. The overall accuracy is impressively high at 97.71%, with a specificity of 96%. In summary, the proposed LSTM-based model outperforms all the other models listed in terms of specificity, accuracy, and sensitivity, making it a highly promising method for the early detection of PD through FOG episode detection. Fig. 6 illustrates the comparative analysis between the proposed LSTM architecture and state-of-the-art algorithms.

## 5   Conclusion

This paper introduces an LSTM network for the early and noninvasive diagnosis of PD based on gait patterns. The LSTM, well-suited for sequential data analysis, demonstrates proficiency in extracting long-term dependencies from time series data. Due to variations in sample sizes across different inputs, we divide the input frame into two segments to normalize it for the LSTM network. The incorporation of dropout as well as L2 regularization techniques effectively mitigates data overfitting, as evidenced by plots of accuracy and loss. The paper addresses PD detection as a binary classification problem, employing an Adam optimizer to classify PD and healthy subjects. Evaluation of the LSTM classifier's performance employs key metrics as precision, sensitivity, specificity, and accuracy, demonstrating superior results compared to SOTA in gait-based PD diagnosis.

The proposed LSTM-based model demonstrates outstanding performance, especially with its remarkable sensitivity of 99%, indicating an extremely low rate of false negatives. Moreover, the overall accuracy is impressively high at 97.71%, coupled with a specificity of 96%. Moreover, the proposed model surpasses existing models in sensitivity, accuracy, and specificity, establishing it as an exceptionally promising approach for the early diagnosis of Parkinson's disease by identifying freezing of gait episodes.